\documentclass[runningheads]{llncs}
\usepackage{cite}
\usepackage[T1]{fontenc}
\usepackage{amsmath}  
\usepackage{booktabs}
\usepackage{amssymb}
\usepackage{makecell}
\usepackage[colorlinks=true,citecolor=blue,urlcolor=black]{hyperref}
\usepackage{marvosym}
\usepackage{multirow}
\usepackage{caption}      
\usepackage{graphicx,verbatim}

\begin{document}

\title{VAP-Diffusion: Enriching Descriptions with MLLMs for Enhanced Medical Image Generation}
\titlerunning{VAP-Diffusion}
%
\author{Peng Huang\inst{1,2} \and
Junhu Fu\inst{1,2}\and
Bowen Guo\inst{1,2}\and
Zeju Li\inst{1,2}\and
Yuanyuan Wang\inst{1,2}\and
Yi Guo\inst{1,2}\textsuperscript{\Letter}
}
\authorrunning{F. Author et al.}
%
\institute{College of Biomedical Engineering, Fudan University, Shanghai 200433, China\and Key Laboratory of Medical Imaging Computing and Computer Assisted Intervention of Shanghai, Shanghai 200032, China\\
\email{guoyi@fudan.edu.cn}\\
\url{https://github.com/YiBaiHP/VAP-Diffusion}
}



\maketitle              
\begin{abstract}

As the appearance of medical images is influenced by multiple underlying factors, generative models require rich attribute information beyond labels to produce realistic and diverse images. For instance, generating an image of skin lesion with specific patterns demands descriptions that go beyond diagnosis, such as shape, size, texture, and color. However, such detailed descriptions are not always accessible. To address this, we explore a framework, termed Visual Attribute Prompts (VAP)-Diffusion, to leverage external knowledge from pre-trained Multi-modal Large Language Models (MLLMs) to improve the quality and diversity of medical image generation. First, to derive descriptions from MLLMs without hallucination, we design a series of prompts following Chain-of-Thoughts for common medical imaging tasks, including dermatologic, colorectal, and chest X-ray images.  Generated descriptions are utilized during training and stored across different categories. During testing, descriptions are randomly retrieved from the corresponding category for inference. Moreover, to make the generator robust to unseen combination of descriptions at the test time, we propose a Prototype Condition Mechanism that restricts test embeddings to be similar to those from training. Experiments on three common types of medical imaging across four datasets verify the effectiveness of VAP-Diffusion.


\end{abstract}
\section{Introduction and Motivation}

The rapid advancement of deep learning has enabled its clinical applications \cite{dayarathna2024deep}. However, collecting large annotated medical datasets remains challenging due to high costs and privacy concerns, limiting the performance of deep-learning-based diagnosis systems. Conditional generative models, such as Generative Adversarial Networks(GANs) and Diffusion Models \cite{goodfellow2020generative, ho2020denoising}, have been explored for data sharing \cite{han2020breaking, ozbey2023unsupervised}, data augmentation~\cite{ktena2024generative, muller2023multimodal} and causal inference~\cite{huang2024chest}. 

Advanced generative models, such as Diffusion Models~\cite{han2023medgen3d, khader2022medical, bluethgen2024vision}, still fall short in realistic and diverse medical image generation, largely due to the variability of lesions and tissue backgrounds, which stems from the inherent complexity of medical images. Factors like disease status and imaging conditions significantly impact image content, resulting in pronounced distinctions both within and between image classes. Taking skin lesion image generation as an example, even images from the same category (i.e. diagnosed disease) can vary in texture, shape, size, and color (c.f. Fig.~\ref{fig:Comparison_Experiment}). We believe that by providing fine-grained descriptions to the generative model, the model can more effectively capture complex data distributions. \emph{By doing so, the image generation tasks are simplified from matching the density of the entire distribution to matching those of separate individual distributions}. Specifically, in a class-conditioned setting, rather than relying solely on single-class conditions (e.g., disease type), we aim to condition the generated images on enriched attributes (e.g., shape, size).

Since demanded detailed descriptions are often unavailable in practice, we turn to Multi-modal Large Language Models (MLLMs) to generate faithful image descriptions as generation guidance. To this end, we propose Visual Attribute Prompts (VAP)-Diffusion, a framework that leverages external knowledge from MLLMs for enhanced medical image generation. Given the remarkable zero-shot capabilities of MLLMs in general image understanding~\cite{li2022blip} and spatial reasoning~\cite{chen2024spatialvlm}, we believe the knowledge embedded in MLLMs can provide valuable insights to the generative models, thus improving the realism and diversity of generated images.

Deriving valid descriptions from pre-trained MLLMs is non-trivial because MLLMs are likely to produce inaccurate or fabricated information that is not even related with the input images. This phenomenon is usually referred as hallucination~\cite{wei2022chain}. MLLMs are more likely to generate hallucinations when analyzing medical images, as they are predominantly trained on natural images. Therefore, we carefully design a Visual Attribute Prompt Strategy (VAPS) following Chain-of-Thoughts~\cite{wei2022chain}, aiming at eliciting the reasoning potential of MLLMs for medical images. Our method is applicable to a spectrum of common medical images, including dermatologic, colorectal, and chest X-ray images.

While generated descriptions of training samples can be readily utilized for training, corresponding descriptions are unavailable at test time. Therefore, we propose a Class-Specific Prompt Bank (CSPB) in VAP-Diffusion, which stores descriptions from training samples and randomly retrieves class-specific descriptions during inference. Furthermore, we introduce a Prototype Condition Mechanism (PCM) to ensure the test embeddings are similar to those from training phase, enabling the model to generalize well to unseen description combinations. 

In summary, the contributions of VAP-Diffusion are summarized as follows:
\begin{itemize}
    \item We design VAPS for VAP-Diffusion, which prompts the pre-trained MLLMs to generate accurate and informative descriptions as enriched guidance.
    \item We propose CSPB along with PCM to ensure that the image generator can be readily used without descriptions or with out-of-distribution descriptions.
    \item We extensively evaluate VAP-Diffusion on three common medical image datasets and demonstrate that the generated images are more realistic and diverse than those produced by existing algorithms. Moreover, we find that VAP-Diffusion can boost the performance of downstream classification tasks by up to $11.9\%$ compared to its counterpart.
\end{itemize}

\begin{figure*}[!h]
  \centering
   \includegraphics[width=0.9\linewidth]{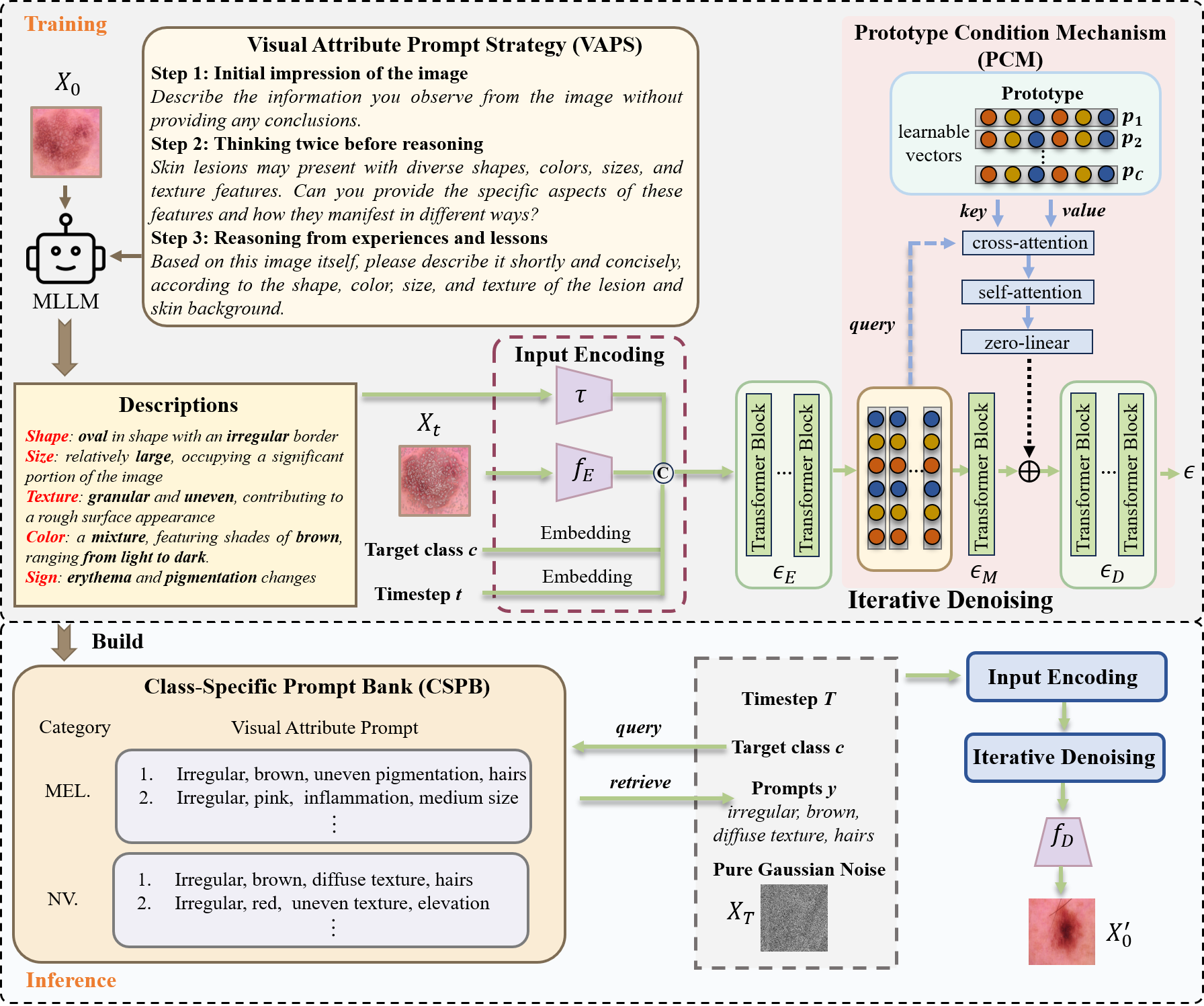}
   \caption{The overview of VAP-Diffusion. Our framework is built upon a diffusion model and consists of three modules: VAPS, CSPB, and PCM. Together, these modules aim to enhance the generator with enriched descriptions from MLLMs and enable it to work with arbitrary descriptions as conditions during testing.
   }
   \label{fig:framework}
\end{figure*}
\section{Methodology}

We aim to develop a diffusion model capable of generating realistic and diverse medical images, thus enhancing the performance and reliability of computer-aided diagnostic models. As shown in Fig.~\ref{fig:framework}, VAP-Diffusion includes three key components: VAPS, CSPB, and PCM, detailed in the following sections.

\subsection{Preliminaries}
\label{method:preliminary}
We build VAP-Diffusion on a state-of-the-art diffusion-based generative model UViT~\cite{bao2023all}. Diffusion model learns the real data distribution by gradually adding noise to an input image $X_0$ and reversing the Markov noising process from purely random noise. During the forward process, for step \textit{t}, we calculate $X_t$ from $X_{t-1}$ using:
\begin{equation}
    q(X_t|X_{t-1}) = \mathcal{N}(X_t|\sqrt{1-\beta_t}X_{t-1},\beta_t \boldsymbol{I}),
    \label{equ1}
\end{equation}
where $\mathcal{N}$ denotes the Gaussian distribution, $\boldsymbol{I}$ denotes the identity matrix, $\beta_t \in (0,1)$ is a step-varying hyper-parameter. Specifically, the cumulative transition from $X_0$ to $X_t$ can be represented as $X_t = \sqrt{\bar{\alpha}_t} X_0 + \sqrt{1 - \bar{\alpha}_t} \epsilon, \epsilon \sim \mathcal{N}(\boldsymbol{0}, \boldsymbol{I})$ where $\bar{\alpha}_t = \prod_{s=1}^t (1 - \beta_s)$ is the cumulative scaling factor. 

Assuming the total number of steps is $T$, the generative model learns to reverse the forward process using following objective function: 
\begin{equation}
    \mathcal{L}_D = \mathbb{E}_{X_0, \epsilon \sim \mathcal{N}(\boldsymbol{0}, \boldsymbol{I}), t}\Big [\|\epsilon-\epsilon_\theta(X_t,t)\|^2_2\Big ],
    \label{equ2}
\end{equation}
where $\epsilon_\theta$ refers to the denoising network. During inference, the trained model $\epsilon_\theta$ progressively denoises the data by generating \(X_{t-1}\) from \(X_t\) for \(t = T, T-1, \dots, 1\) until we reconstruct the image \(X_0'\) which should be similar to the original images. Notably, we use a pre-trained encoder $f_E$~\cite{rombach2022high} to map images to the latent space and reconstruct images using a trainable decoder $f_D$.

\subsection{Visual Attribute Prompt Strategy}
\label{method:visual}
We explore prompt strategy to guide the pre-trained MLLM to generate accurate descriptions based on pre-defined basic visual attributes such as shape, color, size and texture. Here, we take dermatologic image as example to illustrate VAPS and will share VAPS for other modalities (i.e. colorectal and chest X-ray).
\\

\noindent \textbf{Step 1: Initial impression of the image.}
First, we aim to obtain a basic understanding of given image $X_0$. MLLM is directly required to describe the dermatologic image according to template \textit{$Question^1$}, yielding $t^1 = \text{MLLM}(X_0, Question^1)$.

\noindent \emph{$Question^1$:"You are an AI visual assistant observing a skin lesion image. Describe the information you observe from the image without providing any conclusions."}
\\

\noindent \textbf{Step 2: Thinking twice before reasoning.}
Next, we guide MLLMs with \textit{$Question^2$}, obtaining $t^2 = \text{MLLM}(Question^2)$. Based on predefined visual attributes including texture, size, shape, and color, the MLLM is required to consider the inherent imaging characteristics of chosen medical modalities, instead of direct observation of the input images. The generated response remains objective and independent of $t^1$, ensuring a robust and unbiased interpretation.
\\
\\
\textit{$Question^2$: "Skin lesions may present with diverse shapes, colors, sizes, and texture features. The surrounding skin may also appear normal or show signs of inflammation, pigmentation, or other changes. Can you provide the specific aspects of these features and how they manifest in different ways?"}
\\
\\

\noindent \textbf{Step 3: Reasoning from experiences and lessons.}
Finally, to further mitigate hallucinations in the MLLM, we ask the MLLMs to leverage answers from the previous two stages and generate an accurate and concise summary of the given image, resulting in $t^{mix} = \text{MLLM}(t^1, t^2, X_0, Question^3)$.
\\
\\
\textit{$Question^3$: "Based on this image itself, please describe it shortly and concisely, according to the shape, color, size (due to the unavailability of quantitative dimensions, you can just describe their approximate proportion in the whole image), and texture of the lesion and skin background."}
\\

We then encode $t^{mix}$ with BiomedCLIP \cite{zhang2023biomedclip} $\tau$ to ensure compatibility with the image processing pipeline.

\subsection{Class-Specific Prompt Bank}
\label{method:Class_Prompt_Bank}
To supplement descriptions for the inference stage, we build CSPB to store the enriched descriptions for each class during training. Specifically, during training, for each image $X_0$ in the dataset, VAP-Diffusion utilizes a pre-trained MLLM to generate detailed image descriptions $y$. For a given class \textit{c}, we make a collection of $\{y_i\}_{i=1}^n$, where \textit{n} represents the number of samples belonging to class \textit{c}. During inference, for a given class \textit{c}, a prompt $y_i$ is randomly selected from the prompt bank and used as an additional condition to guide image generation. Notably, we also evaluate our model using a visual attribute prompt bank that differs from the one used during training, ensuring that VAP-Diffusion remains flexible enough to handle challenging cases.

\subsection{Prototype Condition Mechanism}
\label{method:Prototype_condition}
To make VAP-Diffusion flexible enough to handle free-text inputs, we propose PCM to enhance the model's robustness against unseen descriptions through prototype matching for each category. Specifically, during the inference stage, given class information and arbitrary descriptions, we aim to regularize the embeddings to remain close to those of training samples from the same class.

Given that the denoising network $\epsilon_\theta$ consists of encoder blocks $\epsilon_E$, middle blocks $\epsilon_M$ and decoder blocks $\epsilon_D$. For each time step \textit{t} and one sample $X_t$, we calculate the encoding features as $F_{e} = \epsilon_E(f_E(X_t), t, c, \tau(y)) \in \mathbb{R}^{K}$. Based on these, prototypes of all $C$ classes are constructed. For each class $c$, we produce $\boldsymbol{p}_c$ which contains $K$ elements. To build the connection between prototypes and input class $c$, $\boldsymbol{p}_c$ is optimized with $\mathcal{L}_{recon}$ to reconstruct $F_{e}$ at any time step $t$ via a cross-attention layer. We further employ a self-attention layer to enhance the multi-modal representation. To stabilize the training process, we build a linear layer initialized with all zero which can gradually inject the multi-modal priors.

With balancing term $\alpha$, our objective function for a single optimization is:
\begin{equation}
    \begin{aligned}
        \mathcal{L}_{VAP}= \mathbb{E}_{X_0, \epsilon \sim \mathcal{N}(\boldsymbol{0}, \boldsymbol{I}), t}\Big [||\epsilon-\epsilon_\theta (f_E(X_t), t, c, \tau(y))||_2^2 \Big ] + \alpha \mathcal{L}_{recon}.
    \end{aligned}
\end{equation}

\begin{table}[!h]  
  \centering  
  \begin{minipage}{0.6\textwidth}  
    \centering  
    \resizebox{\linewidth}{!}{  
      \begin{tabular}{l l c c c c}  
        \toprule  
        \textbf{Dataset} & \textbf{Method} & \textbf{FID $\downarrow$} & \textbf{IS $\uparrow$} & \textbf{Precision $\uparrow$} & \textbf{Recall $\uparrow$} \\
        \midrule  
        \multirow{6}{*}{ISIC2018}  
         & StyleGAN & \textbf{16.968} & 2.967 & \textbf{0.640} & 0.427 \\
         & CBDM & 59.963 & 3.066 & 0.319 & 0.128 \\
         & LDM & 53.375 & 3.491 & 0.395 & 0.264 \\
         & DiT & 57.368 & 2.974 & 0.373 & 0.251 \\
         & UViT & 52.760 & 3.353 & 0.346 & 0.275 \\
         & VAP-Diffusion (Ours) & 19.790 & \textbf{4.404} & 0.617 & \textbf{0.812} \\
        \midrule  
        \multirow{6}{*}{ISIC2019}  
         & StyleGAN & \textbf{23.725} & 4.347 & \textbf{0.457} & 0.360 \\
         & CBDM & 55.676 & 4.798 & 0.329 & 0.254 \\
         & LDM & 57.853 & 4.962 & 0.341 & 0.279 \\
         & DiT & 59.816 & 5.184 & 0.281 & 0.277 \\
         & UViT & 43.292 & 5.113 & 0.372 & 0.304 \\
         & VAP-Diffusion (Ours) & 25.232 & \textbf{5.252} & 0.416 & \textbf{0.496} \\
         \midrule
        \multirow{6}{*}{ChestXray14}   
         & StyleGAN & \textbf{36.221} & 2.296 & 0.343 & 0.418 \\
         & CBDM & 52.139 & 2.419 & 0.276 & 0.416 \\
         & LDM & 47.730 & 2.592 & 0.343 & 0.450 \\
         & DiT & 50.016 & 2.560 & 0.282 & 0.459 \\
         & UViT & 56.204 & 2.723 & 0.295 & 0.455 \\
         & VAP-Diffusion (Ours) & 40.487 & \textbf{2.728} & \textbf{0.375} & \textbf{0.600} \\
        \midrule  
        \multirow{6}{*}{\makecell{Colonoscopy \\ Database}}
         & StyleGAN & 57.871 & 2.699 & 0.214 & 0.341 \\
         & CBDM & 55.692 & 4.725 & 0.334 & 0.386 \\
         & LDM & 46.244 & 4.604 & 0.464 & 0.442 \\
         & DiT & 53.163 & 4.776 & 0.334 & 0.406 \\
         & UViT & 46.265 & 4.779 & 0.395 & 0.492 \\
         & VAP-Diffusion (Ours) & \textbf{32.147} & \textbf{4.812} & \textbf{0.405} & \textbf{0.612} \\
        \bottomrule  
      \end{tabular}  
    }  
    \caption{Quantitative comparison of image synthesis results with other SOTA algorithms.}  
    \label{tab:comparison}  
  \end{minipage}  
  \hfill  
  \begin{minipage}{0.38\textwidth}  
    \centering  
    \resizebox{\linewidth}{!}{  
      \begin{tabular}{l l c c}  
        \toprule  
        \multicolumn{2}{c}{\textbf{Dataset}} & \textbf{FID $\downarrow$} & \textbf{Recall $\uparrow$} \\
        \midrule  
        \multirow{3}{*}{ISIC2018}  
         & VAP-Diffusion & \textbf{26.767} & \textbf{0.609} \\
         & w/o VAPS & 52.760 & 0.275 \\
         & w/o PCM & 28.821 & 0.582 \\
        \midrule  
        \multirow{3}{*}{ISIC2019}  
         & VAP-Diffusion & \textbf{30.357} & \textbf{0.361} \\
         & w/o VAPS & 43.292 & 0.304 \\
         & w/o PCM & 31.931 & 0.338 \\
        \midrule  
        \multirow{3}{*}{ChestXray14}  
         & VAP-Diffusion & \textbf{44.318} & \textbf{0.509} \\
         & w/o VAPS & 56.204 & 0.455 \\
         & w/o PCM & 46.335  & 0.485 \\
        \midrule  
        \multirow{3}{*}{\makecell{Colonoscopy \\ Database}} 
         & VAP-Diffusion & \textbf{32.303} & \textbf{0.519} \\
         & w/o VAPS & 46.265  & 0.492 \\
         & w/o PCM & 37.261 & 0.501 \\
        \bottomrule  
      \end{tabular}  
    }  
    \caption{Ablation experiment results for VAP-Diffusion with unseen input texts. The PCM module only functions with the VAPS module and remains inactive in its absence}  
    \label{tab:Ablation}  
  \end{minipage}  
\end{table}

\begin{table}
  \centering
    \resizebox{0.9\linewidth}{!}{  
      \begin{tabular}{l l c c c c c c}  
        \toprule  
        \textbf{Dataset} & \textbf{Classification Model} & \multicolumn{6}{c}{\textbf{Ratio of Real Images}} \\
        \cmidrule(lr){3-8}  
        & & \multicolumn{3}{c}{\textbf{mAUC}} & \multicolumn{3}{c}{\textbf{F1-score}} \\
        \cmidrule(lr){3-5} \cmidrule(lr){6-8}  
        & & \textbf{1\%} & \textbf{10\%} & \textbf{100\%} & \textbf{1\%} & \textbf{10\%} & \textbf{100\%} \\
        \midrule  
        \multirow{6}{*}{ISIC2018}  
         & Densenet-121 & 0.555 & 0.866 & 0.964 & 0.212 & 0.414 & 0.710 \\
         & + synthetic StyleGAN & 0.913 & 0.913 & 0.965 & 0.526 & 0.576 & 0.741 \\
         & + synthetic VAP-Diffusion (Ours) & \textbf{0.923} & \textbf{0.943} & \textbf{0.978} & \textbf{0.645} & \textbf{0.697} & \textbf{0.753} \\
         & maxViT & 0.555 & 0.867 & 0.964 & 0.412 & 0.413 & 0.713 \\
         & + synthetic StyleGAN & 0.912 & 0.926 & 0.964 & 0.522 & 0.570 & 0.727 \\
         & + synthetic VAP-Diffusion (Ours) & \textbf{0.923} & \textbf{0.940} & \textbf{0.971} & \textbf{0.638} & \textbf{0.687} & \textbf{0.744} \\
         \midrule
        \multirow{6}{*}{ChestXray14 } & Densenet-121  & 0.527 & 0.684 & 0.788 & / & / & / \\
         & + synthetic StyleGAN  & 0.610 & 0.684 & 0.790 & / & / & / \\
         & + synthetic VAP-Diffusion (Ours) & \textbf{0.725} & \textbf{0.766} & \textbf{0.802} & / & / & / \\ & maxViT  & 0.515 & 0.683 & 0.779 & / & / & / \\
         & + synthetic StyleGAN  & 0.607 & 0.684 & 0.784 & / & / & / \\
         & + synthetic VAP-Diffusion (Ours) & \textbf{0.718} & \textbf{0.756} & \textbf{0.799} & / & / & / \\
        \midrule
        \multirow{7}{*}{\makecell{Colonoscopy \\ Database}} & Densenet-121  & 0.583 & 0.879 & 0.923 & 0.272 & 0.803 & 0.907 \\
         & + synthetic StyleGAN  & 0.903 & 0.904 & 0.925 & 0.702 & 0.859 & 0.918 \\
         & + synthetic VAP-Diffusion (Ours) & \textbf{0.923} & \textbf{0.935} & \textbf{0.946} & \textbf{0.744} & \textbf{0.880} & \textbf{0.940} \\ & maxViT  & 0.568 & 0.869 & 0.919 & 0.272 & 0.803 & 0.896 \\
         & + synthetic StyleGAN  & 0.896 & 0.899 & 0.923 & 0.692 & 0.859 & 0.914 \\
         & + synthetic VAP-Diffusion (Ours) & \textbf{0.918} & \textbf{0.931} & \textbf{0.943} & \textbf{0.736} & \textbf{0.879} & \textbf{0.930} \\
        \bottomrule  
      \end{tabular}  
    }  
    \caption{Comparison of downstream task results with StyleGAN. ISIC2019 was not adopted for testing as it does not provide official test set.}  
    \label{tab:downstream_comparison}  
\end{table}

\section{Experiments Results}

\label{sec:Experiments results}
\paragraph{\textbf{Datasets.}}
(1) For dermoscopic modality, we use two public datasets including ISIC2018 dataset \cite{codella2019skin} and ISIC2019 dataset \cite{combalia2019bcn20000}. (2) For chest X-ray images, we choose the ChestXray14~\cite{wang2017chestx}, and the dataset is further processed to include only single-label images. (3) For colonoscopic images, we utilize three publicly available datasets \cite{li2021colonoscopy, misawa2021development, mesejo2016computer}, as well as one private dataset. We split ISIC2018 and ChestXray14 datasets into training and testing sets following official guidelines, while randomly splitting ISIC2019 and colonoscopicd with a ratio of 8:2.

\paragraph{\textbf{Evaluation Metrics}}

We quantify model's performance based on two criteria: realism and diversity. We evaluate realism with FID \cite{heusel2017gans} and Precision \cite{sajjadi2018assessing}, where lower FID and higher Precision indicate better image fidelity. We assess diversity using the IS \cite{salimans2016improved} and Recall \cite{sajjadi2018assessing}, with higher values reflecting greater image diversity. For downstream classification tasks, we employ mAUC and F1-score, where higher scores imply better model performance.

\subsection{Comparison with Other SOTA Methods: Image Synthesis}
We compared VAP-Diffusion with five advanced image generation algorithms (StyleGAN \cite{karras2021alias}, CBDM \cite{qin2023class}, LDM \cite{rombach2022high}, DiT \cite{peebles2023scalable}, and UViT \cite{bao2023all}. As shown in Table~\ref{tab:comparison}, VAP-Diffusion consistently achieves the highest IS and Recall on all datasets. Notably, on the Colonoscopy dataset, VAP-Diffusion achieves the best Recall (0.612) and IS (4.812), while maintaining a competitive FID (32.417), indicating its ability to generate high-quality and diverse images.

\begin{figure*}[h]
  \centering
   \includegraphics[width=1.0\linewidth]{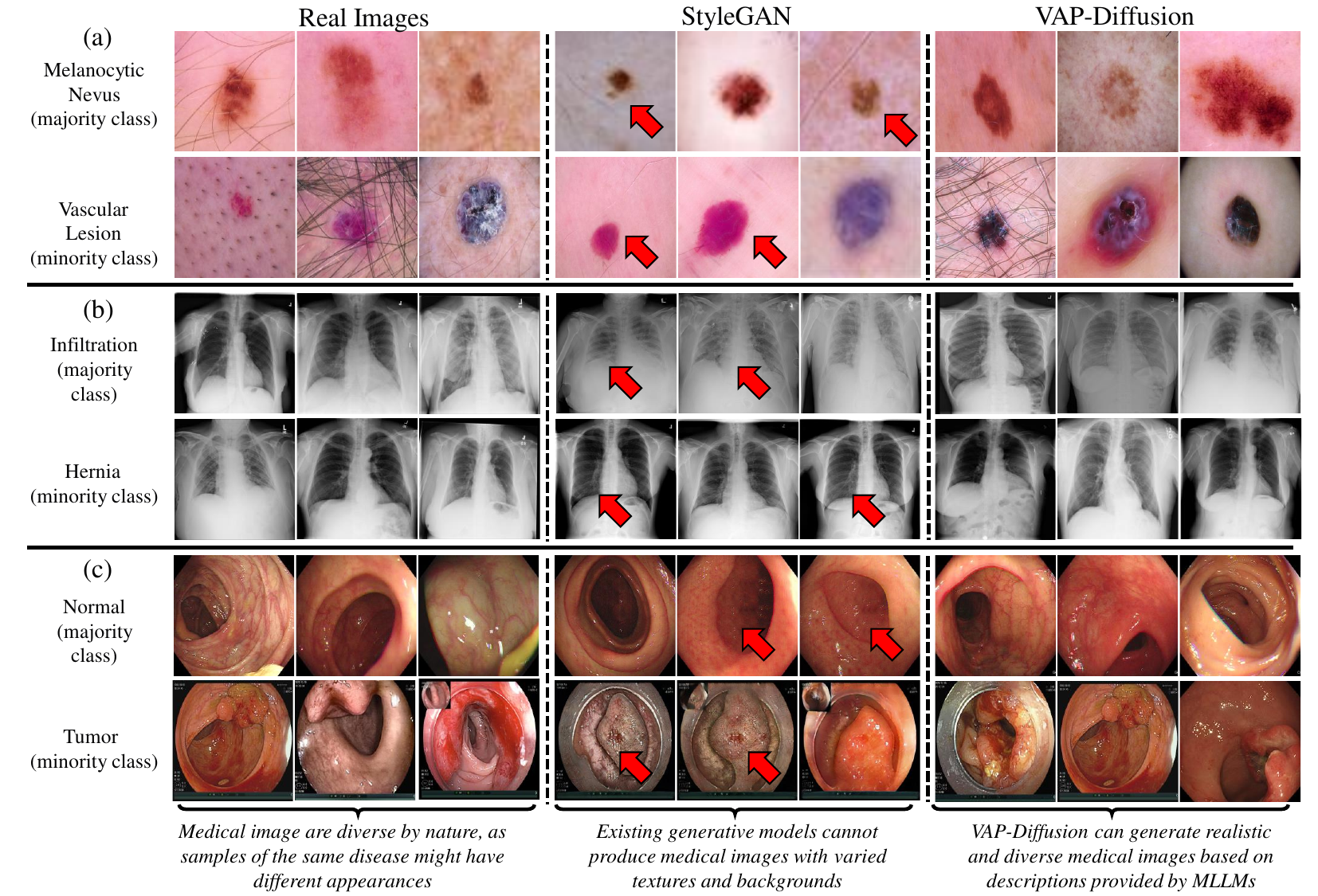}
       \caption{Visual comparison of image synthesis results with StyleGAN. (a) Dermatologic images; (b) Chest X-ray images; (c) Colonoscopic images. Red arrows indicate repetitive content in StyleGAN-generated images, highlighting its potential limitations in producing diverse data.}
   \label{fig:Comparison_Experiment}
\end{figure*}

Additionally, as shown in Fig.~\ref{fig:Comparison_Experiment}, compared to StyleGAN, VAP-Diffusion is capable of generating more realistic and diverse medical images that closely resemble real medical images in terms of texture and structural details. Especially, for dermatologic images, VAP-Diffusion is able to generate complex medical data with visible hairs or rough surface textures.

\subsection{Ablation Experiments}
We conduct ablation experiments on the test sets to evaluate the effectiveness of VAPS and PCM, as well as the robustness of VAP-Diffusion when handling unseen descriptions. As shown in Table \ref{tab:Ablation}, VAP-Diffusion consistently achieves competitive FID and Recall scores across all datasets, even with unseen texts. Of note, on ISIC2018, removing VAPS leads to a significant increase in FID from 26.767 to 52.760 and a sharp decrease in Recall from 0.609 to 0.275, reflecting the importance of VAPS in generating realistic and diverse medical images.

\subsection{Comparison with StyleGAN: Downstream Tasks}

To verify the effectiveness of VAP-Diffusion in downstream tasks, we conducted downstream classification experiments on three modalities using DenseNet \cite{huang2017densely} and maxViT \cite{tu2022maxvit}. As shown in Table~\ref{tab:downstream_comparison}, VAP-Diffusion consistently outperforms both baseline models and those augmented with StyleGAN-generated synthetic data, particularly when only 1\% or 10\% of real data is available. Specifically, on the ISIC2018 dataset with 1\% real data, VAP-Diffusion achieves an mAUC of 0.923 and an F1-score of 0.645, compared to 0.913 and 0.526 for StyleGAN on Densenet-121. This demonstrates the ability of VAP-Diffusion to generate high-quality synthetic data that effectively supplements limited real data. We observe similar improvements on ChestXray14 and the Colonoscopy database.

\section{Conclusion}

\label{sec:Conclusion}
It is difficult for diffusion models to learn complex data distributions in medical scenes with class conditions only. In this paper, we introduce VAP-Diffusion, a novel medical image generative model which successfully exploits the pre-trained MLLM to provide effective guidance for image generation. The key innovation focuses on an efficient Chain-of-Thoughts-based prompt strategy, enabling the MLLM to generate grounded visual attribute prompts. Additionally, we propose a prompt bank along with class prototypes to leverage the multi-modal priors learned during training, assisting diffusion inference. VAP-Diffusion achieves the best generation results on four datasets from three medical modalities and outperforms its counterparts by a large margin on downstream classification tasks.

\begin{credits}
\subsubsection{\ackname} This study was funded by
the National Natural Science Foundation of China (Grant No 62371139), and Shanghai Municipal Education Commission (Grant No. 24KNZNA09).

\subsubsection{\discintname}
The authors have no competing interests to declare that are relevant to the content of this article.
\end{credits}
%
%
%
%





\end{document}